\documentclass[runningheads]{llncs}
\usepackage{graphicx}

\usepackage{tikz}
\usepackage{comment}
\usepackage{amsmath,amssymb} 
\usepackage{color}
 
\usepackage[accsupp]{axessibility}  

\graphicspath{{figures/}}

\def\eg{\emph{e.g.}\,}
\def\ie{\emph{i.e.}\,}

\usepackage{multirow}
\usepackage{booktabs}
\usepackage{arydshln}
\newcommand{\ra}[1]{\renewcommand{\arraystretch}{#1}}
\usepackage{hyperref}
\hypersetup{
	colorlinks=true
}

\begin{document}
\pagestyle{headings}
\mainmatter
\def\ECCVSubNumber{3820}  

\title{Levenshtein OCR} 

\titlerunning{Levenshtein OCR}
%
\author{Cheng Da \thanks{Equal contribution. $\dagger$ Corresponding author.} \and
Peng Wang $^{\star}$ \and
Cong Yao $^{\dagger}$  }

\authorrunning{C. Da et al.}
%
\institute{Alibaba DAMO Academy, Beijing, China\\
\email{\{dc.dacheng08,wdp0072012,yaocong2010\}@gmail.com}}
\maketitle

\begin{abstract}
A novel scene text recognizer based on Vision-Language Transformer (VLT) is presented. Inspired by Levenshtein Transformer in the area of NLP, the proposed method (named Levenshtein OCR, and LevOCR for short) explores an alternative way for automatically transcribing textual content from cropped natural images. Specifically, we cast the problem of scene text recognition as an iterative sequence refinement process. The initial prediction sequence produced by a pure vision model is encoded and fed into a cross-modal transformer to interact and fuse with the visual features, to progressively approximate the ground truth. The refinement process is accomplished via two basic character-level operations:  \textit{deletion} and \textit{insertion}, which are learned with imitation learning and allow for parallel decoding, dynamic length change and good interpretability. The quantitative experiments clearly demonstrate that LevOCR achieves state-of-the-art performances on standard benchmarks and the qualitative analyses verify the effectiveness and advantage of the proposed LevOCR algorithm. Code is available at~\url{https://github.com/AlibabaResearch/AdvancedLiterateMachinery/tree/main/OCR/LevOCR}.
\keywords{Scene Text Recognition, Transformer, Interpretability}
\end{abstract}

\section{Introduction} 
\label{sec:intro}
Scene text recognition is a long-standing and challenging problem~\cite{long2021scene,zhu2016scene,chen2021text} that has attracted much attention from the computer vision community. 
It aims at decoding textual information from natural scene images, which could be very beneficial to down-stream applications, such as traffic sign recognition and content-based image retrieval. However, reading text from natural images is faced with numerous difficulties: variation in text style and shape, non-uniform illumination, partial occlusion, perspective distortion, to name just a few. Recently, various text recognition methods~\cite{long2021scene} have been proposed to tackle this tough problem and substantial progresses have been observed~\cite{DAN,RobustScanner,SRN,ABInet,PIMNet,zhang2022context,he2022visual,liu2022perceiving}.

It has become a trend in the computer vision community to draw inspirations from methods initially proposed for NLP tasks to solve vision problems, for instance, ViT~\cite{dosovitskiy2020image}, DETR~\cite{detr} and Swin-Transformer~\cite{liu2021swin}. Also, in the field of scene text recognition, multiple recent works~\cite{SRN,ABInet} start to incorporate linguistic knowledge into the text recognition process, fusing information from both the vision and language modalities for higher text recognition accuracy.

\begin{figure*}[!htb]\centering
 \includegraphics[width=0.95\textwidth]{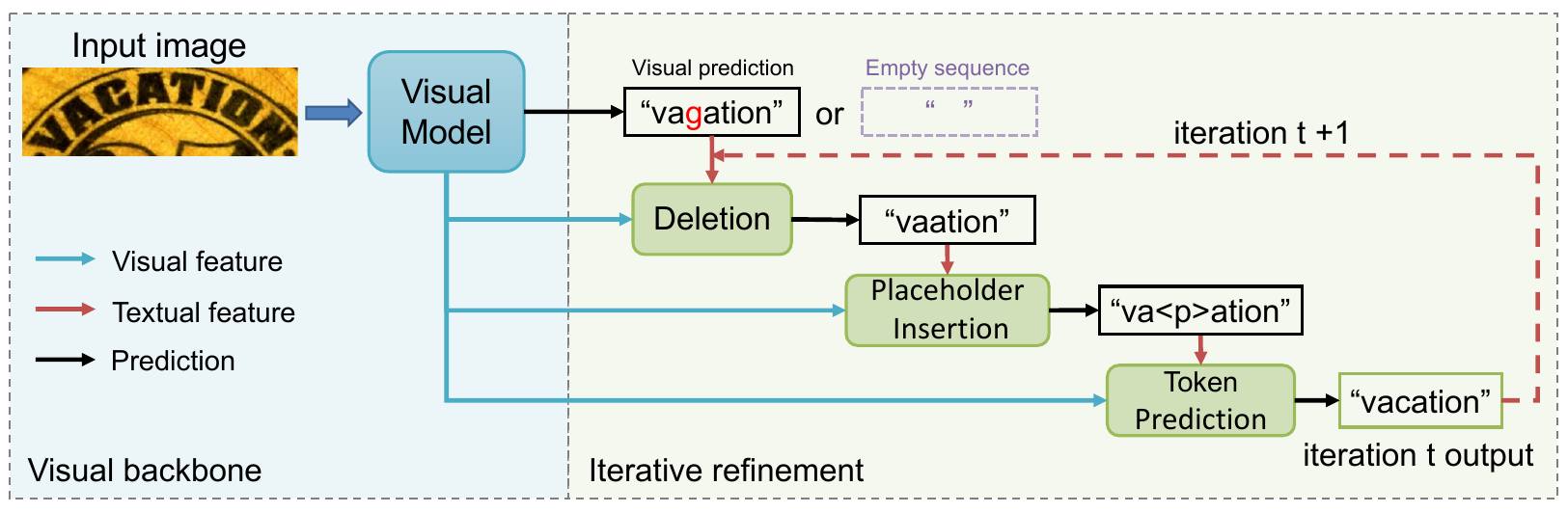}
 \caption{Schematic overview of LevOCR. LevOCR accomplishes text recognition in an iterative way through two basic operations: \textit{deletion} and \textit{insertion}. Note that in LevOCR the operation of \textit{insertion} is further decomposed into two sub-operations: Placeholder Insertion and Token Prediction.}
 \label{fig:motivation}
\end{figure*}

Inspired by the wisdom from these pioneering works, we propose an alternative algorithm for scene text recognition. The backbone of the proposed model is a Vision-Language Transformer (VLT)~\cite{su2019vl,chen2020uniter,VILT}, which is employed to perform cross-modal information fusion for more informative representations and better recognition performance. To further facilitate more flexible and interpretable text recognition, we introduce the strategy from Levenshtein Transformer (LevT)~\cite{levt}, which was originally designed for sequence generation and refinement tasks in NLP, into our framework. The core idea is to learn one refinement policy (deletion or insertion) for current iteration from its adversary in the previous iteration, in order to manipulate the basic units in the sequence (corresponding to characters in text recognition) to approach the target sequence. In such a way, the proposed text recognizer Levenshtein OCR (LevOCR for short) can realize text recognition in a progressive fashion, where the final prediction is obtained by iteratively refining an initial or intermediate recognition result until convergence (\ie, post-editing for error correction). An intuitive illustration is depicted in Fig.~\ref{fig:motivation}. Note that due to the diversity of data augmentation in the training phase, the proposed model also supports generating the final recognition from an \textbf{\textit{empty}} sequence, which falls back to a text generation task.

Similar to ABINet~\cite{ABInet}, we also fuse the information of both the vision modality and the language modality in an iterative procedure to predict the final recognition result. However, there are two key differences: (1) The main architecture of LevOCR is a Vision-Language Transformer (VLT), which allows for more sufficient utilization of the interactions between vision and language; (2) More importantly, while ABINet produces whole sequences at each iteration, LevOCR performs \textbf{\textit{fine-granularity predictions}} through character-level operations (deletion or insertion of individual characters), endowing the system with higher flexibility and better \textbf{\textit{interpretability}}, \ie, when a specific decision (deletion or insertion) is made, one can trace back to the input space (image or text) to examine the supporting cues for that decision. This constitutes an unique characteristic of our LevOCR algorithm.

We have conducted both qualitative and quantitative experiments on widely-used benchmark datasets in the field of scene text recognition to verify the effectiveness of the LevOCR algorithm. LevOCR not only achieves state-of-the-art  recognition performances on various benchmarks (see the tables in Sec.~\ref{Sec:Experiment}), but also provides clear and intuitive interpretation for each action prediction (see the example in Sec.~\ref{Sec:Interpretability} for more details).

In summary, the contributions of the work are as follows: (1) We propose a novel, cross-modal transformer based scene text recognizer, which fully explores the interactions between vision and language modalities and accomplishes text recognition via an iterative process. (2) The proposed LevOCR allows for parallel decoding and dynamic length change, and exhibits good transparency and interpretability in the inference phase, which could be very crucial for diagnosing and improving text recognition models in the future. (3) LevOCR achieves state-of-the-art results on standard benchmarks, and extensive experiments verify the effectiveness and advantage of LevOCR.

\section{Related Work}


\subsection{Scene Text Recognition Methods}
Traditional methods directly cast scene text recognition as a sequence classification task, which is purely based on visual features without any explicitly linguistic knowledge.
CTC-based methods~\cite{CRNN,wan20192d,ctc2,gtc} provide differentiable Connectionist Temporal
Classification (CTC) loss for access to end-to-end trainable sequence recognition,
among which RNN model is often employed for context modeling of feature maps extracted by CNNs\cite{CRNN}.
Segmentation-based methods~\cite{seg,TextScanner} utilize FCN to directly predict the character labels in pixel-level and further group characters into words, in which character-level annotation is required.
Opposite to parallel prediction of CTC and segmentation methods,
attention-based methods~\cite{rnn1,Focusing,ASTER} with encoder-decoder mechanism 
sequentially generates characters in order via RNN-attention model, 
where language information between characters can be implicitly captured.
Due to the promising results, attention-based methods have previously dominated this field.

\subsection{Enhanced Attention-based Methods }
Considering irregular text, previous methods~\cite{ESIR,ASTER} integrate spatial transformer module into attention-based framework, which rectifies the input with perspective and curvature distortion into a more canonical form.
\cite{Focusing} observes the attention drift problem~\cite{Vocabulary},
in which the alignments between feature areas and text targets are not accurate for complicated images,
and proposes a focus network to suppress the  attention adrift.
RobustScanner~\cite{RobustScanner} utilizes positional clues to decode random character sequences effectively, by introducing a position enhancement branch into attention-based framework.
Furthermore, SE-ASTER~\cite{SEED} employs a pre-trained language model to predict additional semantic information, which can guide the decoding process.

\begin{figure}[!htb]\centering
 \includegraphics[width=0.95\textwidth]{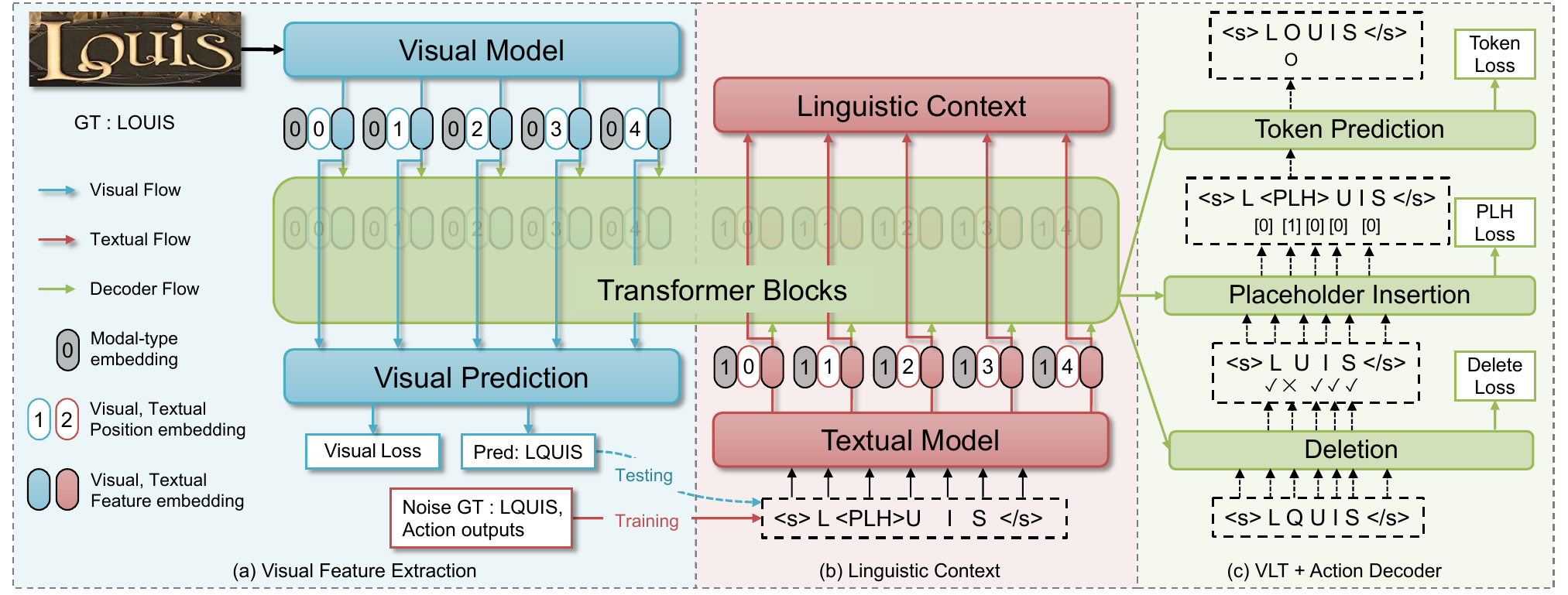}
\caption{Overview of the architecture of LevOCR.}
 \label{fig:overview}
\end{figure}

\subsection{Transformer-based Methods}
Recently, transformer units~\cite{trans} are employed in scene text recognition model to replace the complex LSTM blocks of RNN.
Some prior methods~\cite{SATRN,NRTR,MASTER} provide conventional 
transformer-based encoder-decoder framework for text recognition,
which is based on customized CNNs feature extraction block. 
Furthermore, TrOCR~\cite{TrOCR}  proposes an encoder-decoder structure with pre-trained trans-former-based CV and NLP models, to deal with original image patches directly.
ViTSTR~\cite{ViTSTR} then employs vision transformers (ViT) for scene text recognition with image patches,
where only transformer encoder is required and so that the characters can be predicted in parallel.
To handle the linguistic context well,
VisionLAN~\cite{vlan} presents a visual reasoning module to capture the visual and linguistic information simultaneously
by masking the input image in the feature level.
Additionally, SRN~\cite{SRN} introduces a global semantic reasoning module with transformer units to enhance semantic context.
ABInet~\cite{ABInet} goes a further step and proposes an explicit language model via transformer.

\section{Methodology}

LevOCR consists of three essential modules: Visual Feature Extraction (VFE), 
Linguistic Context (LC) and Vision-Language Transformer (VLT). 
Given an input image, the CNN model first extracts visual information 
and makes an initial visual prediction. 
And the initial visual prediction is further fed into transformer-based textual module to generate linguistic context.
Finally, the intermediate visual features and linguistic features
are directly concatenated and fed into VLT for sufficient information interaction,
without any explicit cross-modal alignments.
Additionally, different action decoders are built upon VLT for subsequent task.
The schematic pipeline  is illustrated  in Fig.~\ref{fig:overview}.

\subsection{Visual Feature Extraction }
Given a image $ \mathbf{x} \in  \mathbb{R}^{H \times W \times 3} $ 
and the corresponding text label $ \mathbf{y} = (y_1,  y_2, \ldots,  y_N)$ with $N$ maximum text length, a modified ResNet~\cite{ASTER,resnet} backbone is utilized for visual information extraction,
and then transformer units~\cite{trans} are employed to generate enhanced 2D visual features $ \mathbf{F}_v \in  \mathbb{R}^{\frac{H}{4} \times \frac{W}{4}  \times D} $, where $ D $ is the feature dimension. 
We directly decrease the height of feature $ \mathbf{F}_v$ to $3$ by $2$ convolution layers,
generating $\mathbf{V} \in  \mathbb{R}^{3 \times \frac{W}{4}  \times D} $ as the visual feature for subsequent refinement task.
In addition, we construct another position attention~\cite{ABInet} branch on feature $ \mathbf{F}_v$ to generate the initial visual prediction $ \hat{\mathbf{y}} = (\hat{y}_1,  \hat{y}_2, \ldots,  \hat{y}_N)$.
Then, the visual loss  $L_{v}$ can be realized by a cross-entropy loss between $ \mathbf{y} $ and $ \hat{\mathbf{y}}$.
Notably, the pure visual feature $\mathbf{V}$ is not used for visual prediction,
in order to preserve more feature information for subsequent refinement task.
The pipeline of visual information extraction is illustrated in Fig.~\ref{fig:overview} (a) with blue color.

\subsection{Linguistic Context}

NLP community has achieved substantial performance improvement. 
to model linguistic knowledge.
The textual module is constructed with Transformer blocks~\cite{trans}
to model linguistic knowledge.
Concretely, the input of textual module is a text sequence that need to be corrected, 
represented by $ \tilde{\mathbf{y}} = (\tilde{y}_1,  \tilde{y}_2, \ldots,  \tilde{y}_N)$. 
First, word embedding at character level is used to encode $ \tilde{\mathbf{y}}$ into feature 
$ \mathbf{F}_t \in  \mathbb{R}^{N \times D} $.
Then, multiple transformer blocks transcribe $ \mathbf{F}_t $ into
refined text feature $\mathbf{T} \in \mathbb{R}^{N \times D }$,
where $N$ is the maximum text length and $ D $ is the feature dimension. 
The pipeline of linguistic context is illustrated in Fig.~\ref{fig:overview} (b) with red color.

\subsection{Vision-Language Transformer}

Text instances in natural scenes do not always conform with linguistic rules. For example, digits and random characters appear commonly.
Therefore, LevOCR employs VLT~\cite{VILT} to integrate visual and linguistic features.
In this way, no enforced alignments as in ABInet~\cite{ABInet} of two modalities are required and adaptive weights of two modalities are directly driven by the objective function.
Then, action decoder heads are built upon VLT for \textit{deletion} and \textit{insertion} action learning, which can make a complementary judgement on both two modalities and be explicitly examined for good interpretability. Specifically, the visual feature $\mathbf{V}=[\mathbf{v}_1; \mathbf{v}_2; \ldots; \mathbf{v}_{N_v} ] \in \mathbb{R}^{N_v \times D } $ 
and textual feature
$\mathbf{T}=[\mathbf{t}_1; \mathbf{t}_2; \ldots; \mathbf{t}_N ]  \in \mathbb{R}^{N \times D } $ are produced by the corresponding modules, respectively.
We directly concatenate these features of two modalities as the input of VLT blocks.
In order to discriminate the features with different modalities and positions, 
position embeddings and modal-type ones are introduced.
The detailed process of VLT is formulated as follows:
\begin{equation}
\begin{split}
&\overline{\mathbf{V}}=[\mathbf{v}_1 + \mathbf{p}^v_1; \mathbf{v}_2 + \mathbf{p}^v_2; \ldots; \mathbf{v}_{N_v} + \mathbf{p}^v_{N_v}] +  \mathbf{E}_v \\
&\overline{\mathbf{T}}=[\mathbf{t}_1 + \mathbf{p}^t_1; \mathbf{t}_2 +  \mathbf{p}^t_2; \ldots; \mathbf{t}_N +  \mathbf{p}^t_N] +  \mathbf{E}_t  \\ 
&\mathbf{H}^{(0)} = [\mathbf{h}^{(0)}_1; \mathbf{h}^{(0)}_2; \ldots; \mathbf{h}^{(0)}_{N + N_v}] =
[\overline{\mathbf{T}}; \overline{\mathbf{V}}] \\
&\mathbf{H}^{(l+1)} = \text{BERT}_l (\mathbf{H}^{(l)} ).
\end{split}
\end{equation}
Here,  $N_v$ is $3 \times \frac{W}{4}$.
The visual and textual position embeddings are represented
as $ [\mathbf{p}^v_1;  \mathbf{p}^v_2; \ldots; \mathbf{p}^v_{N_v}] \in \mathbb{R}^{N_v \times D } $ 
and $ [\mathbf{p}^t_1;  \mathbf{p}^t_2; \ldots; \mathbf{p}^t_N] \in \mathbb{R}^{N \times D } $.  
Then, visual and textual modal-type embeddings are denoted as $  \mathbf{E}_v \in \mathbb{R}^{1 \times D }  $
and $  \mathbf{E}_t  \in \mathbb{R}^{1 \times D } $.
And the $l$-th transformer block is denoted as $ \text{BERT}_l $.
Thus, the final aggregated feature $\mathbf{H} \in  \mathbb{R}^{(N+N_v) \times D } $ are generated by $L$-th transformer block, 
in which even the unaligned features of two modalities can be adaptively interacted and fused.

\subsection{Imitation Learning}
\label{sec:levt}
In order to mimic how humans edit text,  
we cast this text sequence refinement task into a Markov
Decision Process (MDP) denoted as a tuple $(\mathcal{Y,A,E,R}, \mathbf{y}^0)$ as in~\cite{levt}.
We define a text as a sequence that consists of digits and characters, 
and thus $\mathcal{Y}$ is a set of word vocabulary 
with the dictionary of symbols  $ \mathcal{V} $.
Typically, $\mathbf{y}^0 \in \mathcal{Y}$ represents the initial sequence.
For text refinement task, two basic actions $deletion$ and $insertion$ are defined as the set of actions $\mathcal{A}$.
The reward function $\mathcal{R} = - \mathcal{D} (\mathbf{y}, \mathbf{y^*})$ 
directly measures the Levenshtein distance 
between the prediction and the sequence of ground-truth text.	
Given $k$-th step text sequence $\mathbf{y}^k$, 
the agent interacts with the environment $\mathcal{E}$,
executes editing actions and returns the modified sequence $ \mathbf{y}^{k+1}$, 
which is denoted as $ \mathbf{y}^{k+1} = \mathcal{E} (\mathbf{y}^k ,  \mathbf{a}^{k+1} )$.
Our main purpose is to learn a favourable policy $\pi$ that model the probability distribution over actions $\mathcal{A}$ for maximum reward.

\subsubsection{Deletion Action}
The input text sequence for imitation learning  is also denoted as  $ \mathbf{y} = (y_1,  y_2, \ldots,  y_N)$ for simplicity. 
Deletion policy $ \pi^{del}(d | i, \mathbf{y} )$  aims to make a binary decision for every character $y_i \in \mathbf{y}$, in which $d=1$ indicates that this token should be deleted or $d=0$ for keeping it.
Typically, $y_1  $ and $ y_N $ are special symbols $<$s$>$ and $<$/s$>$ for sequence boundary, respectively.
Thus, they can not be deleted, which is denoted as
$ \pi^{del}(0 | 1, \mathbf{y} ) = \pi^{del}(0 | N, \mathbf{y} ) = 1$.
Moreover, based on the aggregated feature  $\mathbf{H}$, 
the deletion classifier can be formulated as follows:
\begin{equation}
\label{eq:del}
 \pi^{del}_\theta(d | i, \mathbf{y} ) = \text{softmax}( \mathbf{h}_i  \mathbf{W}_{del}^T ), i = 2, \dots, N-1,
\end{equation}
where $  \mathbf{W}_{del} \in  \mathbb{R}^{2 \times D } $ is the weight of deletion classifier.
Note that only the first $N$ sequences of $\mathbf{H}$ are used for prediction, and $N$ is the maximum text length.

\subsubsection{Insertion Action}
Insertion action is more complicated than deletion one, since the position of insertion need to be predefined. 
Technically, $insertion$ is decomposed into two sub-operations: 
placeholder insertion and token prediction.
Concretely, for each consecutive pairs $(y_i,  y_{i+1})$ in $\mathbf{y}$,
placeholder insertion policy $ \pi^{plh}(p | i, \mathbf{y} )$ predicts the number  $p$ of 
placeholder should be inserted at position $ i $.
Thus, the classifier of placeholder insertion  is defined as follows:
\begin{equation}
\label{eq:phl}
\begin{split}
 \pi^{plh}_\theta(p | i, \mathbf{y} ) &=  \text{softmax}([\mathbf{h}_i,\mathbf{h}_{i+1}] \mathbf{W}_{plh}^T ), \\
i &= 1, \dots, N-1,
 \end{split}
\end{equation}
where  $  \mathbf{W}_{plh} \in  \mathbb{R}^{M \times 2D }  $ is the weight of the placeholder classifier, 
and $ M $ is the max number of placeholders can be inserted. 
$ [\mathbf{h}_i,\mathbf{h}_{i+1}]$ is the concatenation 
of $\mathbf{h}_i$ and  $\mathbf{h}_{i+1}$.

Referring to the predicted number of placeholder insertion,
we can insert a corresponding number of placeholders at the relevant positions.
Then, token prediction policy $ \pi^{tok}(t | i, \mathbf{y} )$ is required to
replace placeholder $y_i$ with symbol $t$ in the dictionary  $ \mathcal{V} $,
which is formulated as follows:
\begin{equation}
\label{eq:tok}
 \pi^{tok}_\theta(t | i, \mathbf{y} ) = \text{softmax}( \mathbf{h}_i  \mathbf{W}_{tok}^T ), \forall_{y_i}=<\text{p}>.
\end{equation}
Here, $  \mathbf{W}_{tok} \in  \mathbb{R}^{ |\mathcal{V}| \times D }  $ 
is the weight of token predictor, $<$p$>$ is the placehoder and 
$ |\mathcal{V} |$ is the size of dictionary.

\subsubsection{Training Phase}
Notably, $deletion$ and $insertion$ are alternatively executed.
For instance, given a text sequence, deletion policy is first called to delete wrong symbols.
Then, placeholder insertion policy inserts some possible placeholders.
Finally,  token prediction policy replaces all placeholders with right symbols.
Typically, these  actions are performed in parallel for each position.
Moreover, the imitation learning strategy is utilized for LevOCR training, aiming to approximating the expert policy
$ \pi^* $ that can be directly and simply derived from the ground-truth text sequence as follows:
\begin{equation}
\label{eq.optimal}
 \mathbf{a}^*= \mathop{\arg\min}\limits_{\mathbf{a}} \mathcal{D}(\mathbf{y}^*, \mathcal{E} (\mathbf{y}, \mathbf{a})).
\end{equation}
Here, Levenshtein distance $ \mathcal{D} $  is used for distance measure.
The optimal actions $  \mathbf{a}^* $ can represent $\mathbf{d}^*$, $\mathbf{p}^*  $, and $  \mathbf{t}^*  $,
which can be produced by dynamic programming efficiently. The loss function of $deletion$ is formulated as follows:
\begin{equation}
\label{loss:del}
L_{del} = \mathbb{E}_{\mathbf{y}_{del}  \sim d_{{\tilde{\pi}}_{del}}}
\sum_{d^*_i \in \mathbf{d}^*} -\log \pi^{del}_\theta(d^*_i | i, \mathbf{y}_{del} ),
\end{equation}
where $\mathbf{d}^* = (d^*_1, d^*_2, \ldots, d^*_N) \sim \pi^*$ denotes the optimal deletion action for each position of $\mathbf{y}_{del}$,
generated by Eq.~(\ref{eq.optimal}).
And $d_{{\tilde{\pi}}_{del}}$ is a text  distribution induced by policy $\tilde{\pi}_{del}$ 
for the sequence generation with additive  noise:
\begin{equation}
\label{pi:del}
 d_{{\tilde{\pi}}_{del}} = 
 \begin{cases}
 \{\mathcal{E} (\mathcal{E}( \mathbf{y}^0, \tilde{\mathbf{p}}), \tilde{\mathbf{t}} )), 
 \tilde{\mathbf{p}} \sim \pi^{R},  \tilde{\mathbf{t}} \sim \pi^{R}\}, \alpha < \mu \\ 
 \{\mathcal{E} (\mathcal{E}( \mathbf{y}^{\prime}, \mathbf{p}^*), \tilde{\mathbf{t}} )), 
 \mathbf{p}^* \sim \pi^{plh}_{{\theta}_*},  \tilde{\mathbf{t}} \sim \pi^{tok}_{\theta}\},  \alpha \geqslant \mu, \\
\end{cases}
\end{equation}
where $\pi^{R}$ represents a random policy, $\alpha \sim \text{Uniform}[0,1]$,
$ \mu \in [0,1]$ is a mixture factor, $\mathbf{y}^0$ is the initial sequence, 
and $\mathbf{y}^{\prime}$ is any sequence ready to insert.
For $ \mu < \alpha$, we randomly add some symbols on $\mathbf{y}^0$ to generate $\mathbf{y}_{del}$.
For $ \mu \geqslant \alpha$, we use expert placeholder policy
and the learned token prediction to generate  $\mathbf{y}_{del}$ based on $\mathbf{y}^{\prime}$.
This procedure  can be regarded as adversarial learning in GAN~\cite{GAN}. Similarly, the loss function of $insertion$ is  as follows:
\begin{equation}
\label{loss:ins}
\begin{split}
L_{ins} = \mathbb{E}_{\mathbf{y}_{ins}  \sim d_{{\tilde{\pi}}_{ins}}}
[\sum_{p^*_i \in \mathbf{p}^*} - \log \pi^{plh}_\theta(p^*_i | i, \mathbf{y}_{ins} )\\
+\sum_{t^*_i \in \mathbf{t}^*} -\log \pi^{tok}_\theta(t^*_i | i, \mathbf{y}_{ins}^{\prime})],
\end{split}
\end{equation}
where $\mathbf{p}^* = (p^*_1, p^*_2, \ldots, p^*_{N-1}) \sim \pi^* $ represents the optimal number of placeholders
for each consecutive position pair in $\mathbf{y}_{ins} $, generated by Eq.~(\ref{eq.optimal}).
And $\mathbf{t}^* = (t^*_1, t^*_2, \ldots, t^*_{N-1}) \sim \pi^* $ denotes the optimal symbol 
for each placeholder in  $\mathbf{y}_{ins}^{\prime}$, where  $\mathbf{y}_{ins}^{\prime}  = \mathcal{E} (\mathbf{y}_{ins}  ,  \mathbf{p}   )$, $\mathbf{p} \sim  \pi^{plh}_{{\theta}}$.
Moreover, $d_{{\tilde{\pi}}_{ins}}$ is a text  distribution induced by policy $\tilde{\pi}_{ins}$ 
for the sequence generation with deleted noise:
\begin{equation}
\label{pi:ins}
 d_{{\tilde{\pi}}_{ins}} = 
 \begin{cases}
 \{\mathcal{E} (\mathbf{y}^*, \tilde{\mathbf{d}} )), \tilde{\mathbf{d}} \sim \pi^{R}\},  \beta < \mu \\
 \{\mathcal{E} (\mathbf{y}^0, \mathbf{d}^* )), \mathbf{d}^* \sim \pi^{del}_{{\theta}_*}  \},  \beta \geqslant \mu,
\end{cases}
\end{equation}
where factor $\beta \sim \text{Uniform}[0,1]$.
We also adopt mixture manner to construct $\mathbf{y}_{ins}$ for insertion learning.
For $ \beta < \mu$, we randomly delete some symbols on ground-truth $\mathbf{y}^*$ to produce $\mathbf{y}_{ins}$.
For $ \beta \geqslant \mu$, expert deletion policy is employed 
to generate  $\mathbf{y}_{ins}$ based on initial sequence $\mathbf{y}^{0}$. The training procedure is illustrated in Fig.~\ref{fig:overview} (c) with green color, 
and the final loss function is formulated as:
\begin{equation}
\label{loss}
\begin{split}
L = \lambda_1 L_v + \lambda_2 L_{del} +  \lambda_3 L_{ins},
\end{split}
\end{equation}
where $\lambda_1$, $\lambda_2$ and $\lambda_3$ are weights for visual prediction, 
$deletion$ and $insertion$.

Notably, the visual model is pre-trained with only images for better initialization by $ L_v$.
And the textual model and VLT blocks can also be pre-trained with only texts.
Specifically, the input of VLT is $\mathbf{H}^{(0)} = \overline{\mathbf{T}}$ without image feature, which is used for 
 \textit{deletion}  and \textit{insertion}  learning via $ L_{del} $ and  $L_{ins}$.
Based on these pre-trained models, LevOCR is further trained by Eq.~(\ref{loss}).


Note that the input sequences (\ie $\tilde{\mathbf{y}}$) for $deletion$ and $insertion$
are indeed different.
Typically, $\tilde{\mathbf{y}}$ is not always a ``true'' word.
For instance,  $\tilde{\mathbf{y}}$ could be the output of placeholder insertion for token prediction
that includes placeholders or even be an empty sequence.
Therefore, different input sequences $\tilde{\mathbf{y}}$ should be fed into textual model individually 
and encoded as the unique text features $\mathbf{T}$ and aggregated  features $\mathbf{H}$ for the specific action (deletion, placeholder insertion and token prediction) learning in training phase.

\subsection{Inference Phase}
We alternatively employ $deletion$ and $insertion$  to refine text at inference process, until two policies converge
(either nothing to delete or insert, or reaching maximum iterations).
Concretely, given a image $\mathbf{x}$, we first obtain the visual feature $\mathbf{V}$ 
and the initial sequence $ \mathbf{y}^0$ (\eg visual prediction). 
And then $ \mathbf{y}^0$  is fed into textual module, generating refined text texture $\mathbf{T}$.
Furthermore, the visual feature and textual one are interacted and fused in VLT
to produce aggregated feature $\mathbf{H}$.
Finally, action decoders greedily choose the action with the maximum probability
at each position by Eq.~(\ref{eq:del})(\ref{eq:phl})(\ref{eq:tok}) in parallel.
Then, $deletion$ and $insertion$ are executed in turn with new corresponding features $\mathbf{T}$ and $\mathbf{H}$.
Note that LevOCR can not only accomplish text refinement on initial visual predictions,
but also perform text generation with empty sequence  $ \mathbf{y}^0$.

\section{Experiment}
\label{Sec:Experiment}
\subsection{ Datasets}
For fair comparison, we follow previous settings~\cite{deep,ABInet}
to train LevOCR on two synthetic datasets MJSynth (MJ)~\cite{MJ1,MJ2} 
and SynthText(ST)~\cite{ST} without finetuning on other datasets.
Extensive experiments are conducted on six standard Latin scene text benchmarks, 
including 3 regular text datasets (IC13~\cite{IC13}, SVT~\cite{SVT}, IIIT~\cite{IIIT}) and 3 irregular ones (IC15~\cite{IC15}, SVTP~\cite{SVTP}, CUTE~\cite{CUTE}).

\textbf{ICDAR 2013 (IC13)}~\cite{IC13} includes 1095 cropped word images for testing. We evaluate on 857 images with alphanumeric characters and more than $2$ characters.
\textbf{Street View Text (SVT)}~\cite{SVT} contains 647 testing images collected from Google Street View.
\textbf{IIIT 5K-Words (IIIT5k)}~\cite{IIIT} is crawled from Google image search, and consists of 3000 testing images.
\textbf{ICDAR 2015 (IC15)}~\cite{IC15} includes word patches cropped from incidental scene images captured by Google Glasses.
\textbf{Street View Text-Perspective (SVTP)}~\cite{SVTP} consists of 639 images collected from Google Street View, and many images are heavily distorted.
\textbf{CUTE80 (CUTE)}~\cite{CUTE} contains 80 natural scenes images for curved text recognition. 288 images are cropped from these images for testing.

\subsection{Implementation Details}
The textual model and VLT block consist of $6$ stacked transformer units with $8$ heads for each layer, respectively.
The number of hidden units in FC-layer of transformer block is 2048,
and the dimension $D$ of visual and textual feature is set to $512$.
Besides,
The size of symbol dictionary  $ |\mathcal{V} |$ is  $40$,  including $0$-$9$, $a$-$z$, $<$s$>$, $<$/s$>$, $pad$ and  $<$p$>$.
The max length of output sequence $N$ is set to $28$,
and the max number of placeholders $ M $ is set to $28$. 
The mixture factor $ \mu$ is set to $0.5$.
$\lambda_1$, $\lambda_2$ and $\lambda_3$ are set to $1$.
The input images are resized to $32 \times 128$.
Common data augmentation is employed, 
such as rotation, affine, perspective distortion,  image quality deterioration and color jitter.
Our approach is trained on NVIDIA Tesla V100 GPUs with batch size $128$. 
Adadelta~\cite{Adadelta} optimizer is adopted with the initial learning rate $0.1$,
and the max training epoch is $10$.

\begin{table*}[t]\centering
\setlength{\tabcolsep}{4pt}
\ra{1}
\caption{The accuracies of LevOCR with different initial sequences $\mathbf{y}^0$  and max iterations on 6 public benchmarks.}
\label{tab:y0}
\begin{tabular}{|c|c|c|c|c|c|c|c|c|}
\hline
Initial Sequence & Iteration &IC13&SVT  &IIIT   & IC15 & SVTP &CUTE  &AVG \\
\hline
LevOCR$_{VP}$ & - &95.10	&90.57	&95.23	&83.98	&83.41	&87.84	&90.65	 \\
\hline
\multirow{3}{*}{LevOCR + $\mathbf{y}^0_{VP}$} & \#1   &96.50	&92.43	&96.50	&86.09	&87.75	&91.32	&92.55\\
& \#2 &96.73	&92.89	&96.63	&86.42	&88.06	&91.67	&92.79  \\
& \#3 &96.85	&92.89	&96.63	&86.42	&88.06	&91.67	&92.81  \\
\hline
\multirow{3}{*}{LevOCR + $\mathbf{y}^0_{Emp}$} & \#1   &95.45	&90.42	&95.00	&83.99	&83.72	&88.89	&90.64    \\
& \#2  &96.73	&92.43	&96.33	&86.03	&87.60	&92.01	&92.51    \\
& \#3  &96.97	&92.74	&96.37	&86.09	&88.06	&92.01	&92.63   \\
\hline
\multirow{3}{*}{LevOCR + $\mathbf{y}^0_{Rand}$}  & \#1  &84.83	&85.32	&87.90	&82.05	&82.33	&84.38	&85.20 \\
& \#2 &84.83	&85.63	&87.93	&82.05	&82.33	&84.38	&85.25	 \\
& \#3 &84.83	&85.63	&87.93	&82.05	&82.33	&84.38	&85.25  \\
\hline
\multirow{3}{*}{LevOCR + $\mathbf{y}^0_{GT}$} 
& \#1   &99.07	&98.15	&98.67	&92.16	&97.83	&97.22	&96.91 \\
& \#2  	&99.07	&98.15	&98.67	&92.16	&97.83	&97.22	&96.91  \\
& \#3  &99.07	&98.15	&98.67	&92.16	&97.83	&97.22	&96.91  \\
\hline
\end{tabular}
\end{table*}

\subsection{Text Refinement and Text Generation } \label{sec:y0}

\begin{table*}[t]\centering
\setlength{\tabcolsep}{3pt}
\ra{1}
\caption{The accuracies of LevOCR with different backbones.}
\label{tab:vit}
\begin{tabular}{|c|c|c|c|c|c|c|c|c|}
\hline
Methods & Backbone &IC13&SVT  &IIIT   & IC15 & SVTP &CUTE  &AVG \\
\hline
LevOCR$_{VP}$ w/o LevT   &  \multirow{2}{*}{CNN}      &95.21	&90.41	&95.30 &	83.26	&83.41	&88.88	&90.53 \\  
\cline{1-1}
\cline{3-9}
LevOCR w/o LevT  &     &95.21	&90.42	&95.43	&83.43	&84.03	&89.23	&90.70   \\  
\hline
LevOCR$_{VP}$ & \multirow{2}{*}{ViT} &94.86	&89.18	&93.6	&82.38	&84.18	&82.98	&89.30	 \\
\cline{1-1}
\cline{3-9}
LevOCR &  &96.15	&91.80	&95.63	&85.81	&88.06	&86.81	&91.87	 \\
\hline
LevOCR$_{VP}$  &  \multirow{2}{*}{CNN}     &95.10	&90.57	&95.23	&83.98	&83.41	&87.84	& 90.65  \\  
\cline{1-1}
\cline{3-9}
LevOCR   &    &{96.85} &{92.89} &{96.63} &{86.42} &{88.06} &{91.67}    & {92.81} \\
\hline
\end{tabular}
\end{table*}

The initial sequence $\mathbf{y}^0$  is pivotal in LevOCR, 
which not only determines the final performance, 
but also endows LevOCR with the ability of highly flexible text editing.
To empirically verify the ability of text refinement and generation, 
we construct $4$ kinds of initial sequence:
(1) $\mathbf{y}^0_{VP}$ : the visual prediction is directly adopted and LevOCR essentially aims to text refinement.
(2) $\mathbf{y}^0_{Emp}$: empty sequence is simply used and thus the inference stage falls back to a text generation task.
(3) $\mathbf{y}^0_{Rand}$ :  ground-truth is corrupted by random noise, where we replace one character for $30\%$ text, add one character for $30\%$ text, delete one character for $40\%$ text, and remain the digit text unchangeable.
(4) $\mathbf{y}^0_{GT}$ : ground-truth.
These results with different initial sequences are reported in Tab.~\ref{tab:y0}.

\begin{table}[!htb] \centering
\scriptsize
\caption{The detailed iterative process of LevOCR with different initial sequences on 6 public benchmarks. 
``Iteration'' represents the current number of refinement iteration.
``Input'' represents the initial sequence or the output of last iteration. 
``\textcolor{red}{Delete}'', ``\textcolor{cyan}{PLHIns}'' and ``\textcolor{magenta}{TokenPred}''  
represent the action output of deletion, placeholder insertion and token prediction, respectively. 
And \textcolor{cyan}{$<$3p$>$} indicates that three consecutive placeholders $<$p$>$ are inserted into the sequence. Best viewed in colors.}  \label{tab:process}
\renewcommand\arraystretch{1.4}
  \begin{tabular}{ | c | c | l | l |}
    \hline
    Image & GT &Initial  Sequence &Iteration: Input-\textcolor{red}{Delete}-\textcolor{cyan}{PLHIns}-\textcolor{magenta}{TokenPred}\\ 
    \hline
    {$\texttt{1}$: IC13} & \multirow{5}{*}{service} 
    & $\mathbf{y}^0_{VP}$ = servce & \#1: servce-\textcolor{red}{sere}-\textcolor{cyan}{ser$<$3p$>$e}-\textcolor{magenta}{service}\\
    \cline{3-4}
    \multirow{3}{*}{
    \begin{minipage}[b]{0.15\columnwidth}
		\centering
		\raisebox{-.5\height}{\includegraphics[width=\linewidth]{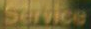}}
	\end{minipage}
	}
    & & \multirow{3}{*}{$\mathbf{y}^0_{Emp}$ = $<$s$>$ $<$ /s $>$} & \#1: $<$s$>$ $<$ /s $>$-\textcolor{red}{$<$s$>$ $<$ /s $>$}-\textcolor{cyan}{$<$7p$>$}-\textcolor{magenta}{serbice}\\
    & & &\#2: serbice-\textcolor{red}{serbice}-\textcolor{cyan}{serb$<$p$>$ice}-\textcolor{magenta}{serblice}\\
    & & &\#3: serblice-\textcolor{red}{serice}-\textcolor{cyan}{ser$<$p$>$ice}-\textcolor{magenta}{service}\\
     \cline{3-4}
    & &$\mathbf{y}^0_{Rand}$ = servcce  &\#1: servcce-\textcolor{red}{servce}-\textcolor{cyan}{serv$<$p$>$ce}-\textcolor{magenta}{service}\\
     \hline
 
    {$\texttt{2}$: SVT}     & \multirow{5}{*}{public} & $\mathbf{y}^0_{VP}$ = publif &\#1: publif-\textcolor{red}{publi}-\textcolor{cyan}{publi$<$p$>$}-\textcolor{magenta}{public}\\
     \cline{3-4}
    \multirow{3}{*}{
    \begin{minipage}[b]{0.15\columnwidth}
		\centering
		\raisebox{-.5\height}{\includegraphics[width=\linewidth]{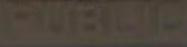}}
	\end{minipage}
	}
    & & \multirow{2}{*}{$\mathbf{y}^0_{Emp}$ = $<$s$>$ $<$ /s $>$} &\#1: $<$s$>$ $<$ /s $>$--\textcolor{red}{$<$s$>$ $<$ /s $>$}-\textcolor{cyan}{$<$6p$>$}-\textcolor{magenta}{publif}\\
    & & &\#2: publif-\textcolor{red}{publi}-\textcolor{cyan}{publi$<$p$>$}-\textcolor{magenta}{public}\\
    \cline{3-4}
    & & \multirow{2}{*}{$\mathbf{y}^0_{Rand}$ = publcc} &\#1: publcc-\textcolor{red}{publc}-\textcolor{cyan}{publc$<$2p$>$}-\textcolor{magenta}{publcip}\\
    & & &\#2: publcip-\textcolor{red}{publi}-\textcolor{cyan}{publi$<$p$>$}-\textcolor{magenta}{public}\\
    \hline
    
    {$\texttt{3}$: IIIT} & \multirow{4}{*}{solaris} & $\mathbf{y}^0_{VP}$ = solris &\#1: solris-\textcolor{red}{soris}-\textcolor{cyan}{so$<$2p$>$ris}-\textcolor{magenta}{solaris}\\
     \cline{3-4}
    \multirow{2}{*}{
    \begin{minipage}[b]{0.15\columnwidth}
		\centering
		\raisebox{-.5\height}{\includegraphics[width=\linewidth]{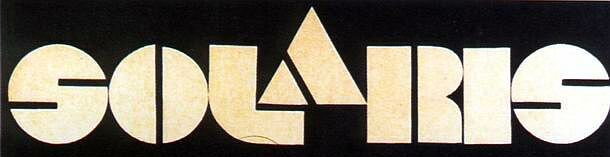}}
	\end{minipage}
	}
    & & \multirow{2}{*}{$\mathbf{y}^0_{Emp}$ = $<$s$>$ $<$ /s $>$} &\#1: $<$s$>$ $<$ /s $>$-\textcolor{red}{$<$s$>$ $<$ /s $>$}-\textcolor{cyan}{$<$6p$>$}-\textcolor{magenta}{solris}\\
    & & &\#2: solris-\textcolor{red}{soris}-\textcolor{cyan}{so$<$2p$>$ris}-\textcolor{magenta}{solaris}\\
     \cline{3-4}
    & &$\mathbf{y}^0_{Rand}$ = solacis  &\#1: solacis-\textcolor{red}{solais}-\textcolor{cyan}{sola$<$p$>$is}-\textcolor{magenta}{solaris}\\
    \hline
    
    {$\texttt{4}$: IC15} & \multirow{4}{*}{breakfast} & $\mathbf{y}^0_{VP}$ = breakeast &\#1: breakeast-\textcolor{red}{breakast}-\textcolor{cyan}{break$<$p$>$ast}-\textcolor{magenta}{breakfast}\\
     \cline{3-4}
    \multirow{2}{*}{
    \begin{minipage}[b]{0.15\columnwidth}
		\centering
		\raisebox{-.5\height}{\includegraphics[width=\linewidth]{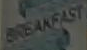}}
	\end{minipage}
	}
    & & \multirow{2}{*}{$\mathbf{y}^0_{Emp}$ = $<$s$>$ $<$ /s $>$} &\#1: $<$s$>$ $<$ /s $>$-\textcolor{red}{$<$s$>$ $<$ /s $>$}-\textcolor{cyan}{$<$9p$>$}-\textcolor{magenta}{breakeast}\\
    & & &\#2: breakeast-\textcolor{red}{breakast}-\textcolor{cyan}{break$<$p$>$ast}-\textcolor{magenta}{breakfast}\\
     \cline{3-4}
    & &$\mathbf{y}^0_{Rand}$ = breaufast  &\#1: breaufast-\textcolor{red}{breafast}-\textcolor{cyan}{brea$<$p$>$fast}-\textcolor{magenta}{breakfast}\\
    \hline
    
   {$\texttt{5}$: SVTP}  & \multirow{6}{*}{house} & $\mathbf{y}^0_{VP}$ = houce &\#1: houce-\textcolor{red}{hou}-\textcolor{cyan}{hou$<$2p$>$}-\textcolor{magenta}{house}\\
     \cline{3-4}
    \multirow{4}{*}{
    \begin{minipage}[b]{0.15\columnwidth}
		\centering
		\raisebox{-.5\height}{\includegraphics[width=\linewidth]{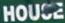}}
	\end{minipage}
	}
    & & \multirow{2}{*}{$\mathbf{y}^0_{Emp}$ = $<$s$>$ $<$ /s $>$} &\#1: $<$s$>$ $<$ /s $>$-\textcolor{red}{$<$s$>$ $<$ /s $>$}-\textcolor{cyan}{$<$5p$>$}-\textcolor{magenta}{houce}\\
    & & &\#2: houce-hou-\textcolor{cyan}{hou$<$2p$>$}-\textcolor{magenta}{house}\\
    \cline{3-4}
    & & \multirow{3}{*}{$\mathbf{y}^0_{Rand}$ = housk} &\#1: housk-\textcolor{red}{houk}-\textcolor{cyan}{houk$<$2p$>$}-\textcolor{magenta}{houke2}\\
    & & &\#2: houke2-\textcolor{red}{hou2}-\textcolor{cyan}{hou$<$p$>$2}-\textcolor{magenta}{hous2}\\
    & & &\#3: hous2-\textcolor{red}{hou}-\textcolor{cyan}{hou$<$2p$>$}-\textcolor{magenta}{house}\\
    \hline
    
    {$\texttt{6}$: CUTE}  & \multirow{3}{*}{vacation} & $\mathbf{y}^0_{VP}$ = vagation &\#1: vagation-\textcolor{red}{vaation}-\textcolor{cyan}{va$<$p$>$ation}-\textcolor{magenta}{vacation}\\
     \cline{3-4}
    \multirow{2}{*}{
    \begin{minipage}[b]{0.15\columnwidth}
		\centering
		\raisebox{-.5\height}{\includegraphics[width=\linewidth]{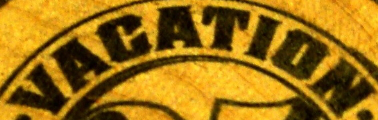}}
	\end{minipage}
	}
    & & $\mathbf{y}^0_{Emp}$ = $<$s$>$ $<$ /s $>$ &\#1: $<$s$>$ $<$ /s $>$-\textcolor{red}{$<$s$>$ $<$ /s $>$}-\textcolor{cyan}{$<$8p$>$}-\textcolor{magenta}{vacation}\\
     \cline{3-4}
    & &$\mathbf{y}^0_{Rand}$ = vacction  &\#1: vacction-\textcolor{red}{vaction}-\textcolor{cyan}{vac$<$p$>$tion}-\textcolor{magenta}{vacation}\\
    \hline
  \end{tabular}
\end{table}

The performance of $\mathbf{y}^0_{Rand}$ and $\mathbf{y}^0_{GT}$ can be regarded as the lower bound and upper bound of LevOCR, respectively.
Based on the noisy initial sequences, $ \mathbf{y}^0_{Rand} $ gets $85.25\%$ average accuracy.
This implies that almost $85\%$ noisy texts can be corrected.
Moreover, $\mathbf{y}^0_{GT}$  achieves the best accuracy $96.91\%$ than others,
proving that better linguistic context can be captured in better initial text sequence, leading to better performance of LevOCR. Additionally, the first line LevOCR$_{VP}$ records the accuracy of pure visual prediction of LevOCR.
Both $\mathbf{y}^0_{VP}$ and $\mathbf{y}^0_{Emp}$ can obtain obvious improvements over LevOCR$_{VP}$, showing the effectiveness of text refinement and generation of LevOCR. Generally, the accuracies of $4$ constructions  
(\ie $\mathbf{y}^0_{GT} > \mathbf{y}^0_{VP}  > \mathbf{y}^0_{Emp} > \mathbf{y}^0_{Rand}$)
suggest that choosing a better $\mathbf{y}^0$ is crucial,
and LevOCR indeed adopts linguistic information, not just relies on visual features.
These results sufficiently demonstrate the powerful abilities of LevOCR in text refinement and text generation.

\subsection{Iterative Refinement}

We investigate the influence of various maximum refinement iterations (\ie, $1, 2, 3$) and the results are also shown in Tab.~\ref{tab:y0}.
Notably, even if the maximum iteration is set to $3$, the refinement process might be stopped after $2$ iterations 
when the sequence remains the same as the first iteration.
Thus the results may be the same under different maximum iterations.
Apparently, the results at \#$1$ are already promising, 
and more iterations lead to further improvements.
Therefore, there is no sophisticated setting for the maximum iteration and the adaptive convergence enables access to high efficiency.
Thus, the maximum iteration is set to $3$ for convenience.
It is crucial to note that the accuracy of $\mathbf{y}^0_{Emp}$ at \#$1$ is close to LevOCR$_{VP}$, which is only based on visual feature.
While the accuracy of $\mathbf{y}^0_{Emp}$  is greatly increased $2.1\%$ at \#$2$ iteration over \#$1$, 
demonstrating that LevOCR indeed utilizes language knowledge for further text refinement.

\subsection{ Effectiveness of Levenshtein Transformer}  \label{Sec:Generation}

To further explore the significance of Levenshtein Transformer pipeline for text refinement,
we construct some variants of LevOCR and these results are shown in Tab.~\ref{tab:vit}.
We remove the \textit{deletion} and \textit{insertion} losses in~Eq.(\ref{loss}) and replace them with only one classification loss, the performance of LevOCR without LevT is not significantly improved and approximate to pure image prediction.
While the proposed LevOCR with LevT obtains obvious $2.4\%$ improvement than pure visual prediction. 
Besides CNN feature, vision transformer (ViT)~\cite{dosovitskiy2020image} with $4 \times 4$ patches is utilized for image feature extraction. And the proposed \textit{deletion} and \textit{insertion} actions are learned for text refinement.
ViT-based LevOCR  can also achieve stable improvement $2.9\%$ than visual prediction.
These results demonstrate that LevT is a perfect fit for text refinement  and the proposed LevOCR works well with CNN and ViT backbones, resulting in promising generalization.


\subsection{Qualitative Analyses} \label{Sec:Qualitative}

It is essential to qualitatively dissect the iterative process of LevOCR.
We select $6$ exemplar images for qualitative exhibition, of which the visual predictions are incorrect. 
Generally, these challenging images are with various types of noises, 
such as motion blur, Gaussian blur, irregular font, occlusion, curved shape, perspective distortion and low resolution.
The iterative refinement processes are elaborately reported in Tab.~\ref{tab:process}.
And different $3$ initial sequences as in Sec.~\ref{sec:y0} are adopted
to show the amending capability powered by $deletion$ and $insertion$.

Deletion action targets at removing wrong letters. As in Img.$\texttt{6}$, 
the `C' is recognized as `G'  by visual model due to the curved shape.
By leveraging both visual and linguistic information, `G' in $\mathbf{y}^0_{VP}$ tends to be deleted.
Similarly, deletion is triggered in $\mathbf{y}^0_{Rand}$ to remove a redundant `C'.
Considering placeholder insertion in empty sequence $\mathbf{y}^0_{Emp}$,
we clearly observe that the inserted number of placeholder equals to the length of GT, 
indicating that LevOCR can fully comprehend image.
When the length of input text  is shorter than GT, placeholder insertion try to add the rest of placeholders at the right position.
Consequently, placeholder insertion and deletion endows LevOCR with the ability of directly altering sequence length, which is different from previous methods (such as SRN~\cite{SRN} and ABINet~\cite{ABInet}).

\begin{table}[!htb]
\setlength{\tabcolsep}{3pt}
\centering
\caption{The accuracy comparisons with SOTA methods on $6$ public benchmarks.
The underlined and bold values represent the second and the best results, respectively.}
\label{tab:sota}
\begin{tabular}{lcccccccc}
\toprule[1.1pt]
\multirow{2}{*}{Methods} & \multirow{2}{*}{Datasets}   &\multicolumn{3}{c}{Regular Text} &\multicolumn{3}{c}{Irregular Text} &\multirow{2}{*}{Average}\\
\cmidrule(lr){3-5}
\cmidrule(lr){6-8}
& &IC13 &SVT  &IIIT   & IC15 & SVTP &CUTE   \\
\midrule[0.8pt]
TBRA~\cite{deep}  &MJ+ST  &93.6 &87.5 &87.9 &77.6 &79.2  & 74.0  & 84.6 \\
ViTSTR~\cite{ViTSTR}  &MJ+ST  &93.2 &87.7 &88.4 &78.5 &81.8  & 81.3  & 85.6 \\
ESIR~\cite{ESIR} &MJ+ST  &91.3  &90.2 &93.3 &76.9 &79.6 &83.3  &  87.1  \\
SAM~\cite{TextSpotter}   &MJ+ST  &95.3 &90.6 &93.9 &77.3 &82.2 &87.8  &  88.3  \\
SE-ASTER~\cite{SEED}   &MJ+ST  &92.8 &89.6 &93.8 &80.0 &81.4 &83.6  & 88.3  \\
TextScanner~\cite{TextScanner}   &MJ+ST  &92.9 &90.1 &93.9 &79.4 &84.3 &83.3  & 88.5 \\
DAN~\cite{DAN}  &MJ+ST  &93.9  &89.2 &94.3  &74.5  &80.0 &84.4   &  87.2  \\
ScRN~\cite{ScRN} &MJ+ST &93.9  & 88.9   &94.4    &78.7 & 80.8   & 87.5 & 88.4 \\
RobustScanner~\cite{RobustScanner}    &MJ+ST   & 94.8 & 88.1  & 95.3 & 77.1 & 79.5 &  90.3 & 88.4 \\
PIMNet~\cite{PIMNet}   &MJ+ST   & 95.2 & 91.2  & 95.2  &83.5  &84.3   &84.4 & 90.5 \\
\hline
SATRN~\cite{SATRN}  &MJ+ST  &94.1 &91.3 &92.8  &79.0 &86.5 &87.8 &  88.6 \\
MASTER~\cite{MASTER}	&MJ+ST &95.3	&90.6	&95.0	&79.4	&84.5	&87.5 & 89.5 \\
\hline
SRN~\cite{SRN}    &MJ+ST   & 95.5 & 91.5  & 94.8  &82.7  &85.1   &87.8 & 90.4 \\
ABINet~\cite{ABInet}      &MJ+ST   &\textbf{97.4} &\textbf{93.5} &\underline{96.2} &\underline{86.0} &\textbf{89.3} &89.2  & \underline{92.6} \\
\hline
LevOCR   &MJ+ST    &\underline{96.85} &\underline{92.89} &\textbf{96.63} &\textbf{86.42} &\underline{88.06} &\textbf{91.67}    & \textbf{92.81} \\
\bottomrule[1.1pt]
\end{tabular}
\end{table}

As for token prediction, it heavily relies on the output of placeholder insertion.
When both predicted placeholders and the rest characters are correct,
token prediction can directly make right decision (such as Img.$\texttt{4}$ and Img.$\texttt{6}$).
However, when the output of deletion or placeholder insertion is incorrect, the refinement might collapse.
As shown in Img.$\texttt{5}$ with $\mathbf{y}^0_{Rand}$, `k' is not removed and $2$ more placeholders are inserted.
Fortunately, Img.$\texttt{5}$ can be successfully corrected after $2$ more iterations.
This phenomenon demonstrates that $deletion$ and $insertion$ are well learned by imitation learning, and adversarial learning guarantees that these two actions are complementary and inter-inhibitive.
Additionally, token prediction supports inserting not only one character, but also word piece (multiple characters), such as ``vic'' in Img.$\texttt{1}$, ``la'' in Img.$\texttt{3}$ and ``se'' in Img.$\texttt{5}$. 
These results clearly confirm the superiority of our method.

\subsection{Comparisons with State-of-the-Arts}

We compare our LevOCR against thirteen state-of-the-art scene text recognition methods on $6$ public benchmarks, and the recognition results are illustrated in Tab.~\ref{tab:sota}.
For fair comparison, we only choose the methods that trained on synthetic datasets MJ and ST, 
and no lexicon is employed for evaluation.
Specifically, LevOCR outperforms SATRN~\cite{SATRN} and MASTER~\cite{MASTER} that are transformer-based encoder-decoder models, showing the effectiveness of BERT-based framework for text recognition.
In addition, SRN~\cite{SRN} and ABINet~\cite{ABInet} achieve impressive performance
by explicitly modelling linguistic information in their methods.
The proposed LevOCR gains $2.4\%$ improvement
over SRN on average accuracy.
Meanwhile, LevOCR surpasses the performance of ABINet on IIIT, IC15 and CUTE datasets
and thus achieves the state-of-the-art performance on average accuracy.
These results demonstrate the effectiveness of LevOCR.

\subsection{Interpretability of LevOCR} \label{Sec:Interpretability}

\begin{figure*}[t]\centering
 \includegraphics[width=0.9\textwidth]{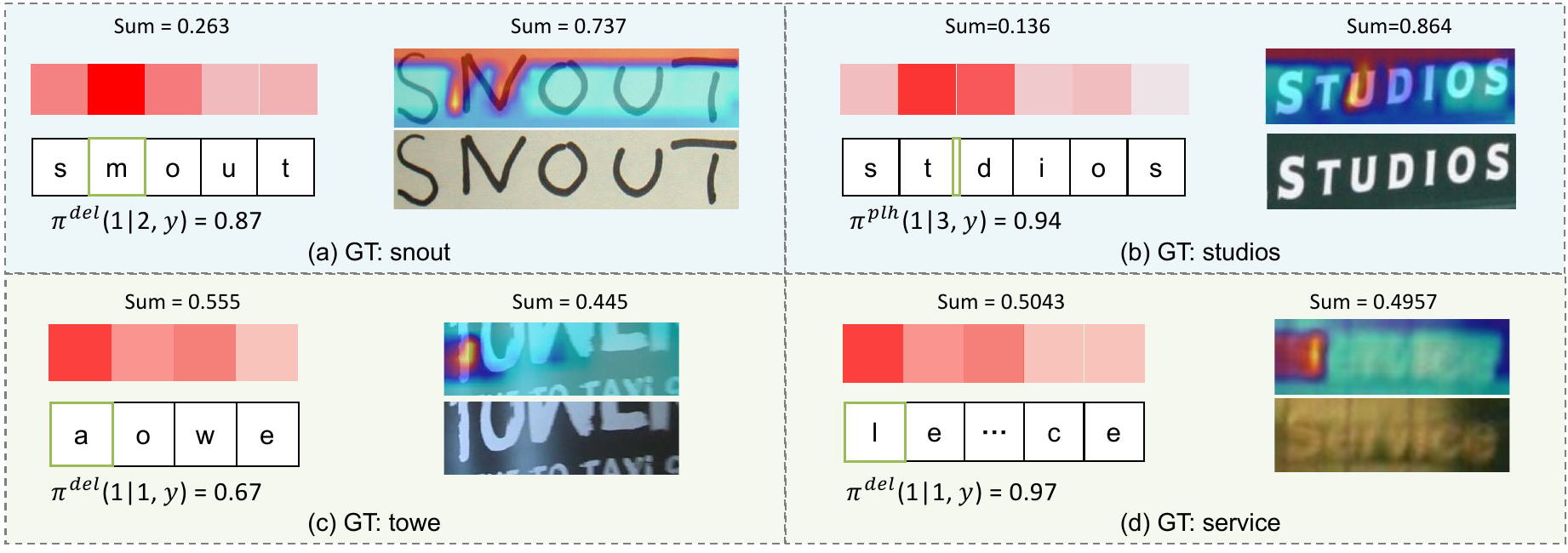}
 \caption{Illustration of the interpretability of LevOCR. The details (attention maps and intermediate results) of the prediction of LevOCR are depicted. The sum attention weights of different modalities in the top layer are visualized. Best viewed in colors.}
 \label{fig:vis}
\end{figure*}

Four intuitive examples to demonstrate the good interpretability of LevOCR are shown in Fig.~\ref{fig:vis}.  As in Fig.~\ref{fig:vis} (a), the second character ``n"  is mis-classified as ``m" in the image with GT string ``snout". LevOCR is able to identify that `m' is wrong and thus should be \textbf{\textit{deleted}} (the probability of deletion is 0.87). Beyond that, one can easily examine the reason why LevOCR makes this decision through attention visualization and quantitative comparison. In this case, LevOCR relies more on visual information than on textual information (0.737~\textit{vs.} 0.263) and the pixels near `n' in the image contribute the most for this decision. Similarly, in the case of Fig.~\ref{fig:vis} (b), where the GT string is ``studios" and `u' is intentionally removed, LevOCR manages to perceive that the third character is missing and thus should be \textbf{\textit{brought back}} (the probability of inserting one character between `t' and `d' is 0.94). 

Additionally, when the initial prediction is incorrect and the image is severely corrupted or blurry, LevOCR will pay more attention to textual information in decision making, as shown in Fig.~\ref{fig:vis} (c) and (d). Through the visualization and numbers produced by LevOCR, we can intuitively comprehend the character-level supporting evidences used by LevOCR in giving this prediction. Notably, good interpretability is an unique characteristic that distinguish the proposed LevOCR from other scene text recognition algorithms. Based on it, we can understand the underlying reason for a specific decision of LevOCR and diagnose its defects. We could even gain insights for designing better systems in the future.

\subsection{Limitation}

The $deletion$ and $insertion$  operations in LevOCR are relatively time-consuming,
where one refinement iteration costs about $36$ ms. Considering the time of visual feature extraction ($11$ ms) and alternative process, the elapsed time is about $47/83/119$ ms for $1/2/3$ iteration. The model size ($109 \times 10^6$ parameters) is relatively large.
Therefore, a more efficient architecture will be our future work.

\section{Conclusion}
In this paper, we have presented an effective and explainable algorithm LevOCR for scene text recognition. Based on Vision-Language Transformer (VLT), LevOCR is able to sufficiently exploit the information from both the vision modality and the language modality to make decisions in an iterative fashion. LevOCR can give fine-granularity (character-level) predictions and exhibits a special property of good interpretability. Extensive experiments verified the effectiveness and advantage of the proposed LevOCR algorithm.


\bibliographystyle{splncs04}
\bibliography{egbib}
\end{document}